\documentclass[a4paper,conference]{IEEEtran}
\IEEEoverridecommandlockouts

\usepackage[numbers,sort&compress,sectionbib]{natbib}
\usepackage{amsmath,amssymb}
\usepackage{graphicx}
\usepackage{booktabs}
\usepackage{multirow}
\usepackage{url}
\usepackage{xcolor}
\usepackage{balance}
\usepackage{array}
\usepackage{hyperref}
\hypersetup{
    colorlinks   = true,
    citecolor    = blue,
    urlcolor    =  blue
}
\usepackage{subcaption}
\usepackage{ragged2e}
\usepackage{tabularx}

\usepackage{listings}

\usepackage{array}
\newcolumntype{L}{X}
\newcolumntype{C}{>{\centering\arraybackslash}X}
\newcolumntype{R}{>{\raggedleft\arraybackslash}X}

\title{\texttt{Math2Visual-X}: A Modular Framework for Pedagogically Aligned Lower-Primary Math Visuals Generation}

\author{
\IEEEauthorblockN{
H.D.E. Maduranga\IEEEauthorrefmark{1},
S. K. Munasinghe\IEEEauthorrefmark{1},
K. P. T. I. Weerasekara\IEEEauthorrefmark{1},
Surangika Ranathunga\IEEEauthorrefmark{2},
Nisansa de Silva\IEEEauthorrefmark{1}
}
\IEEEauthorblockA{\IEEEauthorrefmark{1}
Dept. of Computer Science \& Engineering, University of Moratuwa, Sri Lanka\\
 \{eshan.21, shashini.21, imakula.21\}@cse.mrt.ac.lk, NisansaDdS@cse.mrt.ac.lk
}
\IEEEauthorblockA{\IEEEauthorrefmark{2}
School of Mathematical and Computational Sciences, Massey University, Auckland, New Zealand\\
s.ranathunga@massey.ac.nz
}
}

\begin{document}
\maketitle

\begin{abstract}
Visual representations can help lower-primary learners understand Math Word Problems, but generating classroom-usable visuals remains difficult. Existing symbolic systems are controllable but limited in coverage, while end-to-end text-to-image systems often fail to satisfy exact mathematical constraints. This paper presents a symbolic visual generation framework for lower-primary MWP generation with broader problem coverage and more scalable asset generation. The framework includes an LLM-based routing layer, three worksheet-oriented generation modules, and two fallback mechanisms for open-world SVG asset acquisition. A human evaluation comparing \texttt{Math2Visual-X} with \texttt{Stable Diffusion XL}, \texttt{Nano Banana}, and \texttt{GPT Image} showed that the proposed method achieved the strongest overall performance. The results indicate that the framework offers a scalable and pedagogically grounded approach for automatic MWP visual generation.
\end{abstract}

\begin{IEEEkeywords}
Math Word Problems, educational technology, SVG generation, text-to-image, pedagogical visualization, lower-primary mathematics, Vision Language Models
\end{IEEEkeywords}

\section{Introduction}
Math Word Problems (MWPs) are a long-established part of school mathematics because they connect mathematical operations to meaningful everyday contexts~\cite{hoogland2016representing}. However, lower-primary learners often find MWPs difficult because successful performance draws on text comprehension and oral language~\cite{fuchs2018text}, while also requiring learners to coordinate linguistic and numerical information within limited cognitive capacity~\cite{vessonen2024task}. Therefore, it is common to use MWPs with visual elements with respect to this learner group. Prior work shows that the quality of visual-schematic support matters: accurate representations are associated with substantially better word-problem solving~\cite{boonen2014visual}. Despite extensive work on Math reasoning,  pedagogically appropriate visual generation for lower-primary learners remains relatively underexplored~\cite{wang2025generating}.

Such visual generation is non-trivial. For young learners, a classroom-usable visual should remain semantically faithful to the problem~\cite{wang2025generating}, while avoiding extraneous pictorial detail that can impose unnecessary cognitive load~\cite{vanlieshout2018pictorial}. Even small mismatches in quantities or grouping can disrupt the intended reasoning path, because the usefulness of a representation depends on its structural correctness, rather than on visual plausibility alone~\cite{boonen2014visual, wang2025generating}. These make lower-primary math-visual generation a constrained pedagogical design problem rather than a purely aesthetic image-generation task~\cite{wang2025generating}.

Recent approaches for Math-visual generation can be grouped into two broad families: symbolic composition systems and end-to-end text-to-image systems. 
Symbolic composition systems, such as \texttt{Math2Visual}~\cite{wang2025generating}, transform text into structured specifications and deterministically compose diagrams from reusable assets, providing strong control over correctness while remaining constrained by supported problem types and local asset availability. 
End-to-end text-to-image systems offer wider visual coverage and naturalness, but they remain unreliable for exact object counting~\cite{guo2025text}, attribute binding and compositional reasoning~\cite{huang2023t2icompbench}, and spatial relations~\cite{huang2023t2icompbench,zhang2025compass}, all of which are critical for educational visuals.

This paper presents a Math-visual generation framework (hereafter referred to as \texttt{Math2Visual-X})  in the form of a symbolic composition system. It extends the \texttt{Math2Visual} system, the latest and the best symbolic composition system to date, by addressing its two key limitations: reliance on a fixed asset set and limited operation coverage. Beyond the arithmetic MWPs supported in \texttt{Math2Visual}, \texttt{Math2Visual-X} supports three visually grounded lower-primary MWP types: pattern recognition, object counting, and object attribute comparison. These categories require different visual structures, since pattern tasks emphasize repeated motifs~\cite{rittlejohnson2019patterning}, counting tasks require discrete countable object sets~\cite{clements2021earlymath}, and attribute-comparison tasks rely on salient perceptual contrasts such as size, length, or height~\cite{clements2021earlymath}.


In addition, \texttt{Math2Visual-X} includes an LLM-based routing framework that dispatches MWPs to specialized generation modules, as well as two open-world asset fallback mechanisms that acquire or synthesize missing single-object SVGs when a required icon is unavailable. We also introduce an evaluation protocol to evaluate the quality of the generated Math-visuals. It combines automatic asset-level screening with a completed human evaluation study using a pedagogically grounded rubric. Human evaluation showed that Math2Visual-X outperformed three visual generation models (SDXL, Nano Banana, and GPT Image) across accuracy, completeness, clarity, and cognitive load.


\section{Related Work}
\subsection{Pedagogically Grounded MWP Visualization}
 Recent SVG-generation methods have improved general-purpose vector synthesis. \texttt{SVGDreamer}~\cite{xing2024svgdreamer} explores diffusion-guided text-to-SVG optimization with improved editability, visual quality, and diversity. \texttt{Chat2SVG}~\cite{wu2025chat2svg} combines LLM-based SVG template generation with image-diffusion-based refinement. \texttt{LLM4SVG}~\cite{xing2025llm4svg} improves SVG understanding and generation by adapting LLMs with learnable semantic tokens and SVG-text instruction data.These studies show that SVG generation can benefit from structured language guidance and refinement-based strategies.

However, these methods target general text-to-SVG generation rather than lower-primary mathematical word problems. They mainly emphasize visual fidelity, editability, and broader SVG expressiveness, but do not explicitly model pedagogical requirements such as exact cardinality, clear grouping, operation-faithful composition, and worksheet-style clarity \cite{wu2025chat2svg,
xing2025llm4svg}. For this reason, \texttt{Math2Visual} \cite{wang2025generating} is a more suitable baseline for our work. Unlike general SVG generators, it treats MWP visualization as a structured pedagogical task, mapping problems into a visual language representation and rendering them through controlled SVG composition with explicit entities, quantities, containers, and mathematical relations \cite{wang2025generating}. 

\subsection{Limitations of End-to-End Generative Models}
Direct text-to-image generation remains unreliable for lower-primary educational visuals~\cite{wang2025generating}. Diffusion-based models struggle with exact object counting~\cite{guo2025text}, attribute binding and compositional relations~\cite{huang2023t2icompbench}, and spatial reasoning~\cite{huang2023t2icompbench,zhang2025compass}, even when the generated outputs appear visually plausible. This gap between plausible appearance and pedagogical correctness is critical in classroom visuals, where a single missing object or swapped attribute can change the intended reasoning path. Additionally, these models are stochastic, meaning that the same prompt may produce outputs with different levels of correctness, clutter, or layout stability.

Large vision-language models are also imperfect automatic judges. They hallucinate details inconsistent with an image, and recent work shows that benchmark quality substantially affects measured hallucination behavior \cite{bai2024hallucination,yan2024hallbench}. In addition, they remain weak on visual arithmetic tasks such as counting and comparison \cite{huang2025visualarithmetic,choudhury2025visualequations}. These limitations motivate hybrid approaches that retain explicit structure and use generative methods only where symbolic methods lack coverage.

\subsection{Open-World Asset Creation and Screening}
When reusable visual assets are missing, single-object generation followed by background removal and vectorization becomes a practical option. Latent diffusion models such as \texttt{LDM}~\cite{rombach2022ldm} and \texttt{SDXL}~\cite{podell2024sdxl} provide strong open-vocabulary image synthesis. \texttt{U$^2$-Net}~\cite{qin2020u2net} and practical toolchains such as \texttt{rembg}\footnote{D. Gatis, ``rembg: Remove image backgrounds,'' GitHub repository, 2026. [Online]. Available: \url{https://github.com/danielgatis/rembg}} support foreground extraction and transparent cutout creation. Raster-to-vector tools such as \texttt{VTracer}\footnote{\textit{visioncortex}, ``VTracer: Raster to vector graphics converter,'' GitHub repository, 2020. [Online]. Available: \url{https://github.com/visioncortex/vtracer}} and \texttt{Potrace}~\cite{selinger2003potrace} then enable downstream SVG reuse.

Because generated candidates can still violate educational requirements, automatic screening metrics are useful as triage signals rather than final correctness guarantees. \texttt{CLIP} \cite{radford2021clip} and \texttt{CLIPScore} \cite{hessel2021clipscore} measure prompt-image alignment, \texttt{BLIP} \cite{li2022blip} and \texttt{BLIP-2} \cite{li2023blip2} provide captioning and image-text checking capabilities, and \texttt{ImageReward} \cite{xu2023imagereward}, \texttt{TIFA} \cite{hu2023tifa}, and \texttt{GenEval} \cite{ghosh2023geneval} offer complementary signals for human preference, fine-grained faithfulness, and object-level constraint satisfaction. In this pipeline, these metrics are used as preliminary filters to reject candidates with weak prompt alignment, poor visual quality, or failed instance-level faithfulness checks before the asset is reused in SVG composition.

\section{Methodology}

\begin{figure}[t]
    \centering
    \includegraphics[width=0.8\columnwidth]{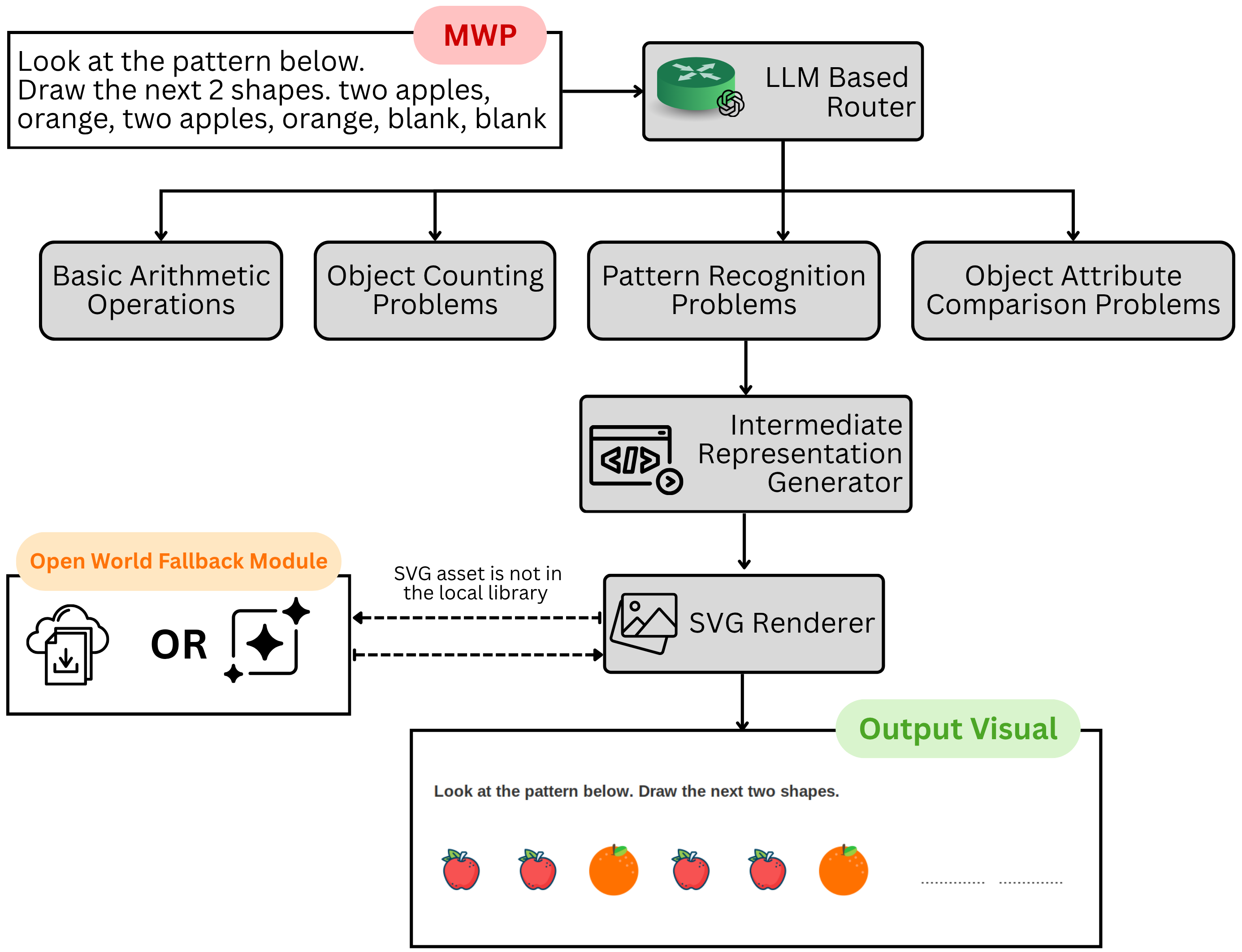}
    \caption{Overall workflow of the framework. }
    \label{fig:system_workflow}
\end{figure}

\begin{figure}[t]
    \centering
    \includegraphics[width=0.8\columnwidth]{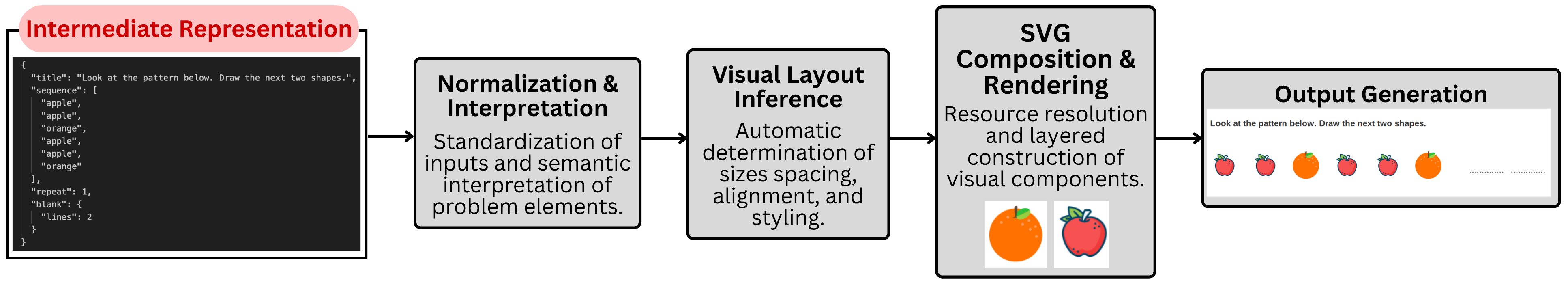}
    \caption{The internal rendering pipeline.}
    \label{fig:rendering_process}
\end{figure}

\begin{table*}
\tiny
\caption{Supported operation types with example prompts and generated visuals.}
\label{tab:categories}
\centering
\renewcommand{\arraystretch}{1.6}
\setlength{\tabcolsep}{6pt}
\small
\resizebox{\textwidth}{!}{
\begin{tabular}{|p{1.8cm}|p{4.3cm}|p{4.3cm}|p{4.3cm}|p{4.3cm}|}
\hline
\textbf{Row} & \textbf{Arithmetic Visual Generation} & \textbf{Pattern Visual Generation} & \textbf{Object Counting Visual Generation} & \textbf{Object Attribute Comparison Generation} \\
\hline

\textbf{Description} 
& Captures entities and operations as an operation tree. The renderer translates this into a layout with operator symbols and grouping cues, compacting crowded scenes while preserving explicit arithmetic structure.
& Converts prompts into sequence payloads defining repetition behaviors. To aid repetition rule inference, the renderer enforces strict consistency in motif size, alignment, and spacing within a single-row layout.
& Transforms prompts into categorical counts, rendered as bounded, mixed-object scenes. Grid layouts prevent overlap for large quantities while intentionally avoiding accidental grouping cues that disrupt direct enumeration.
& Extracts target attributes and relative scales to create highly contrastive layouts. It emphasizes relevant spatial dimensions (e.g., shared baselines for height) while minimizing distractions to ensure unambiguous perceptual judgments. \\
\hline

\textbf{Prompt}
&  Janet has nine oranges, and Sharon has seven
oranges. How many oranges do Janet and Sharon have together?
& Generate a pattern of ball and apple. Repeat 2 times. Use 2 answer line. Title: ``Look at the pattern below and fill in the blanks.''
& Generate an image of ten apples and take ``Count the items.'' as the title.
& Circle the smallest dolphin among the following 4 dolphins with measurements 47, 85, 63, 25. \\
\hline

\textbf{Image}
& \includegraphics[width=0.25\textwidth]{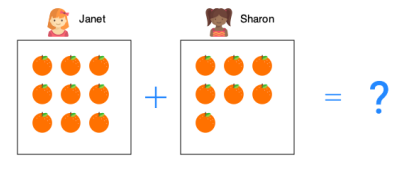}
& \includegraphics[width=0.26\textwidth]{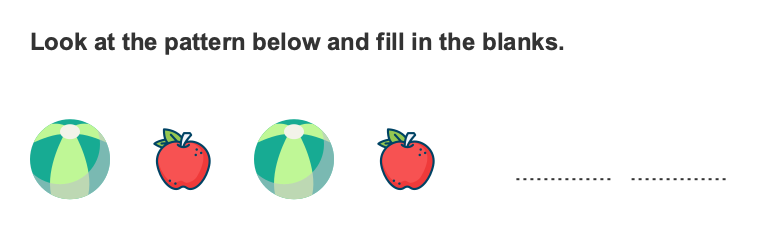}
& \includegraphics[width=0.21\textwidth]{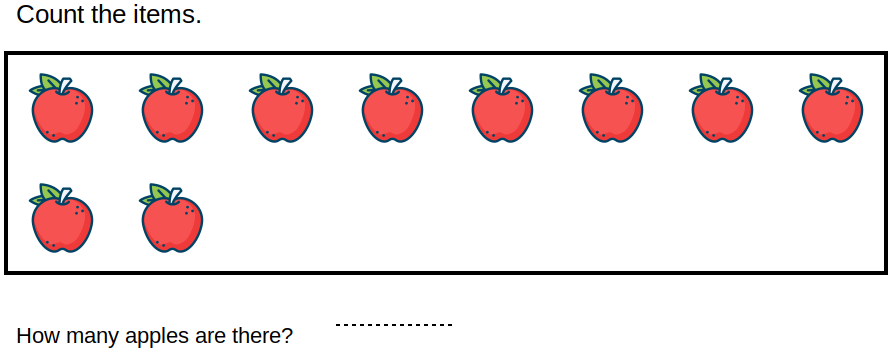}
& \includegraphics[width=0.22\textwidth]{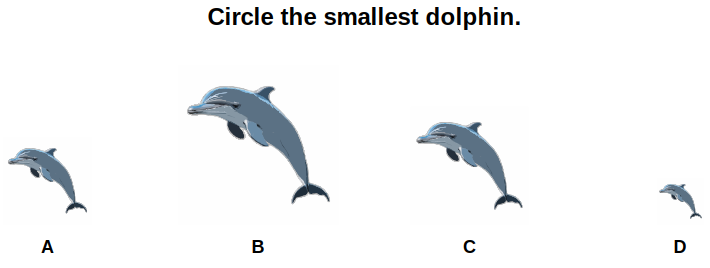} \\
\hline

\end{tabular}}
\end{table*}

In this section, we describe the \texttt{Math2Visual-X} framework, which extends \texttt{Math2Visual}, by supporting a broader set of problem types, including pattern recognition, object counting, and attribute comparison, and by introducing an open-world asset fallback mechanism that enables robust visual generation beyond a fixed asset library. The system architecture is shown in Figure~\ref{fig:system_workflow}.

Although \texttt{Math2Visual-X} supports a diverse range of problem categories, all prompts are processed through a unified modular workflow. A user-provided MWP is first classified by an LLM-based router, for which we implemented and evaluated \texttt{llama3.1:8b} and \texttt{gpt-oss:20b}. Based on the predicted category, the system generates a structured intermediate representation tailored to the semantic requirements of that task, which is then passed to a deterministic SVG renderer (Figure~\ref{fig:rendering_process}). If a required object asset is not available in the local library, an open-world fallback module retrieves or synthesizes a reusable SVG before final composition, and the asset is then embedded into the final image. This architecture enables all prompts to follow a common general pipeline while preserving task-specific visual constraints.

\subsection{Extending the Set of Supported Mathematical Operations}
One central contribution of this work is the extension of symbolic MWP visual generation of \texttt{Math2Visual} beyond arithmetic-centric templates to support three additional lower-primary problem types: pattern recognition, object counting, and object attribute comparison. 
Building on this modular architecture, the framework supports four distinct problem categories. Each category adapts the intermediate representation and rendering logic to meet its specific pedagogical requirements. These new operation types and example prompts are given in Table~\ref{tab:categories}, along with the expected visuals.

\subsection{Open-World Asset Fallback}

\begin{figure}[htbp]
     \centering
     \begin{subfigure}[b]{\columnwidth}
         \centering
         \includegraphics[width=0.8\columnwidth]{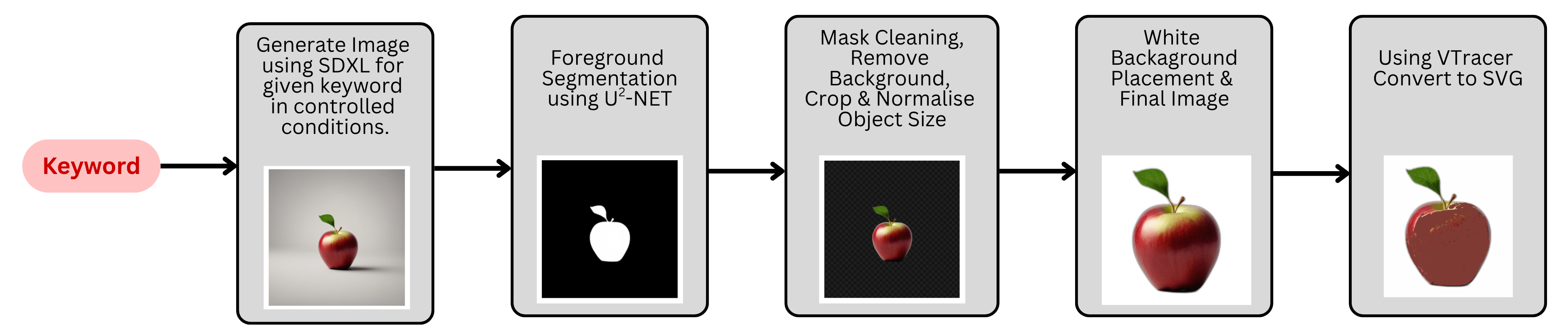}
         \caption{Primary generative synthesis pipeline.}
         \label{fig:svg_generation}
     \end{subfigure}
     
     \vspace{0.3cm}

     \begin{subfigure}[b]{\columnwidth}
         \centering
         \includegraphics[width=0.8\columnwidth]{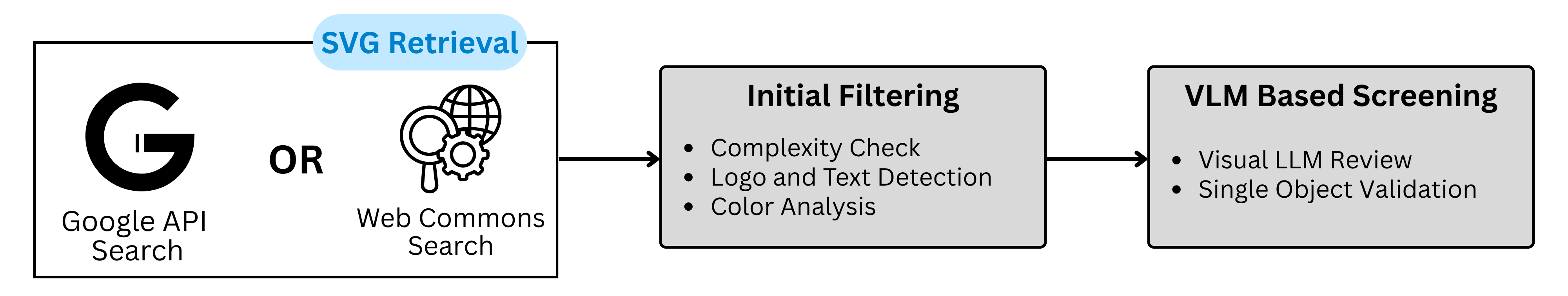}
         \caption{Secondary web-agent retrieval pipeline.}
         \label{fig:retrieval}
     \end{subfigure}
     
     \caption{Open-world asset fallback strategies: (a) prioritized generative synthesis; (b) secondary web-agent retrieval.}
     \label{fig:fallback_strategies}
\end{figure}

As illustrated in Figure~\ref{fig:fallback_strategies}, missing object icons prevent successful symbolic composition. To achieve open-world coverage without relying on a fixed local repository, the framework implements a dual-route fallback mechanism. Based on evaluation results favoring visual coherence, the system prioritizes generative synthesis, falling back to web retrieval only when necessary.

\subsubsection{Generative Single-Object Synthesis}
In the generative strategy, the \texttt{SDXL} latent diffusion model \cite{podell2024sdxl} is used to synthesize high-quality candidate images (see Figure~\ref{fig:svg_generation}). Since diffusion models are optimized for visual richness rather than pedagogical clarity, prompt-level constraints are enforced to generate single-object outputs with clean white backgrounds. The generated image is then processed using \texttt{U$^2$-Net} \cite{qin2020u2net} for accurate salient object detection and foreground extraction. The extracted object is subsequently cropped using its bounding box, with slight padding applied to preserve fine structural details, and then converted into a scalable, reusable SVG via raster-to-vector transformation \footnote{\textit{visioncortex}, ``VTracer: Raster to vector graphics converter,'' GitHub repository, 2020. [Online]. Available: \url{https://github.com/visioncortex/vtracer}}. This approach reliably yields clean, worksheet-friendly assets.

\subsubsection{Web-Agent Retrieval and Curation} 
This strategy activates a web-agent to retrieve clipart-style SVGs (see Figure~\ref{fig:retrieval}). Candidates undergo a heuristic complexity penalty to filter out cluttered scenes, followed by Vision-Language Model (VLM) verification to ensure the asset depicts exactly one centered instance of the target object.

\section{Evaluation Protocol}
To comprehensively evaluate the performance of \texttt{Math2Visual-X}, the evaluation is divided into two main phases: a component-wise evaluation of the system's internal modules to ensure structural correctness, followed by a rigorous human evaluation to measure practical pedagogical effectiveness.

\subsection{Component-Wise Evaluation}
Component-wise evaluation aims to ensure the correctness of each major module in \texttt{Math2Visual-X}. Accordingly, the three key components: the LLM Router, the Intermediate Representation Generation Module, and the Open-World Asset Fallback Mechanism were evaluated independently.

The LLM Router was evaluated based on its ability in classifying and assigning an appropriate tag to a given problem. The Intermediate Representation Generation Module was assessed based on its ability to generate structurally valid and semantically consistent representations aligned with the LLM prompt, which is dynamically selected based on the assigned tag and the given problem. For the Open-World Asset Fallback Mechanism, a combination of model-based evaluation techniques was utilized. Specifically, \texttt{CLIP} evaluates semantic alignment \cite{radford2021clip}, \texttt{ImageReward} assesses visual quality \cite{xu2023imagereward}, and \texttt{TIFA} verifies instance-level faithfulness \cite{hu2023tifa}. 

\subsection{Baseline Systems}
\texttt{Math2Visual-X} was evaluated against three alternative generation systems: \texttt{SDXL} (Stable Diffusion XL), Gemini (\texttt{Nano Banana}), and ChatGPT (\texttt{GPT Image}). The original \texttt{Math2Visual} was excluded because it supports only arithmetic MWPs, whereas our evaluation mainly includes pattern recognition, object counting, and object attribute comparison tasks, making a direct comparison over the full benchmark infeasible. A prompt dataset covering all supported categories was used to generate visuals under the same evaluation setup. The outputs were assessed by primary school teachers and trainee teachers. Thus, \texttt{Math2Visual-X} was evaluated against strong generative baselines using identical prompts and evaluators, with emphasis on pedagogical usefulness rather than visual appeal.


\subsection{Human Evaluation Setup \& Evaluation Rubric}
The evaluation dataset consisted of 100 prompts (400 generated images) covering all four supported problem categories, and the generated outputs were assessed by primary school teachers and teaching students.
The human evaluation rubric was designed after reviewing \texttt{Math2Visual} which introduces the four criteria of Accuracy, Completeness, Clarity, and Cognitive Load. Since \texttt{Math2Visual} did not provide any guideline on these criteria, we defined discrete levels for each criterion (Table~\ref{tab:rubric}). To further ground the rubric, we analyzed lower primary mathematics question papers to identify common visual patterns and structures. Accuracy and Completeness are evaluated against the intended prompt meaning, while Clarity and Cognitive Load are assessed based on the visual output itself. As distinctions between these categories can be subtle, evaluating them from two perspectives improves consistency in human judgments.

\begin{table*}[t]
\tiny
\caption{Human evaluation rubric used for assessing generated math visuals.}
\label{tab:rubric}
\centering
\scriptsize
\setlength{\tabcolsep}{3pt}
\begin{tabularx}{\textwidth}{lLLLLL}
\toprule
\textbf{Criterion} & \textbf{1 -- Very Poor} & \textbf{2 -- Poor} & \textbf{3 -- Fair} & \textbf{4 -- Good} & \textbf{5 -- Excellent} \\
\midrule
Accuracy &
The visual has no recognizable correspondence to the mathematical facts of the problem. Quantities and relationships are absent or completely wrong. &
Multiple key quantities or relationships are incorrect. The visual fundamentally misrepresents the mathematical structure of the problem. &
An inaccuracy is present (e.g., a core quantity is wrong). The mathematical relationship is present but represented incorrectly. &
A minor inaccuracy is present that does not hinder problem-solving, or a non-critical quantity is wrong. &
All quantities, entities, and mathematical relationships are represented perfectly, with no errors. \\
\midrule
Completeness &
The visual is effectively empty or misses almost all essential elements, making it useless for solving the problem. &
Multiple essential elements are missing. The visual only shows a fraction of the problem. &
A key element is missing or an attribute is wrong, making the problem difficult to solve using the visual alone. &
One minor, non-essential element is missing or implied rather than shown. Solvable. &
Contains all essential elements. Every entity, quantity, and relationship from the text is explicitly visualized. \\
\midrule
Clarity &
The visual is incomprehensible, abstract, or fails to render a coherent scene. Impossible to interpret. &
The visual is highly ambiguous. It is difficult to tell what the entities are or how they relate. &
The visual is confusing. Entities are muddled, overlap, or are hard to count. Layout is disorganized. &
The visual is mostly clear, but may require a moment of study to interpret. &
The visual is immediately and unambiguously understandable. Entities are distinct and well organized. \\
\midrule
Cognitive Load &
The visual is overwhelmingly cluttered. Focus is entirely on irrelevant or nonsensical details, obscuring the problem. &
Visual is cluttered with irrelevant details, making it difficult to find essential information. &
Contains distracting elements that draw attention away from the problem core components. &
Includes minor decorative elements that do not contribute to the solution but are not distracting. &
The visual is clean and focused. Includes only elements necessary to solve the problem. \\
\bottomrule
\end{tabularx}
\end{table*}

\section{Results and Discussion}
\subsection{Human Evaluation Results}
To process the evaluation data, the resulting criterion-wise scores were aggregated to obtain per-model averages for Accuracy, Completeness, Clarity, and Cognitive Load, together with an overall average. This makes the comparison consistent across all four systems and ensures that the reported results reflect the same evaluation conditions.

Human evaluation results are presented in Table~\ref{tab:human-results}. \texttt{Math2Visual-X} obtained the best mean scores across all four rubric dimensions: Accuracy, Completeness, Clarity, and Cognitive Load. In contrast, \texttt{SDXL} received the lowest scores across all criteria, while \texttt{Nano Banana} and \texttt{GPT Image} performed better than \texttt{SDXL} but still remained below \texttt{Math2Visual-X}.

\begin{table}[htbp]
\tiny
\caption{Aggregated human evaluation results.}
\label{tab:human-results}
\centering
\footnotesize
\setlength{\tabcolsep}{3pt}
\begin{tabular}{lccccc}
\toprule
\textbf{Model} & \textbf{Acc.} & \textbf{Comp.} & \textbf{Clar.} & \textbf{Cog.} & \textbf{Overall} \\
\midrule
SDXL      & 1.08 & 1.12 & 1.31 & 1.26 & 1.19 \\
Gemini    & 3.74 & 3.85 & 3.97 & 3.87 & 3.86 \\
GPT       & 3.50 & 3.63 & 3.96 & 3.87 & 3.74 \\
Framework & 4.82 & 4.79 & 4.82 & 4.87 & 4.82 \\
\bottomrule
\end{tabular}
\end{table}

The strongest gains appear in Accuracy and Completeness, suggesting that deterministic symbolic generation is better able to preserve the intended entities, quantities, and relationships than general-purpose image generators. The high Clarity and Cognitive Load scores further indicate that the framework produces more worksheet-friendly visuals with less distracting detail, which is important for lower-primary learners.

\textbf{Inter-rater reliability and ranking consistency}
Human evaluation demonstrated strong reliability at both the score and rank levels. Both inter-rater agreement analyses were computed using the same set of 100 common prompts and their corresponding generated images, which were evaluated by all raters. The Inter-rater reliability measured using \textit{ordinal Krippendorff's alpha} yielded an overall coefficient of $\alpha = 0.7457$, indicating acceptable agreement among evaluators and supporting the reliability of the proposed 1--5 rubric for assessing generated educational visuals \cite{krippendorff2011computing,ku2024imagenhub}. To further assess whether evaluators were consistent in the relative ordering of the compared models, we computed prompt-wise Kendall's coefficient of concordance ($W$) \cite{kendall1939}. Across 100 prompts, the \textit{mean prompt-wise $W$ }was \textbf{0.8878}, indicating very strong agreement in evaluator rankings. Taken together, these findings suggest that the evaluation framework was robust, with evaluators showing both reasonably consistent rubric-based judgments and highly consistent comparative ranking behavior across prompts.

\subsection{LLM Router Evaluation Results}
Table~\ref{tab:router-eval-results} presents the F1-scores for the LLM router evaluation, comparing the routing accuracy of \texttt{llama3.1:8b} against \texttt{gpt-oss:20b} on a dataset of 583 questions. To assess the models' ability to predict the correct tag (operation category), the evaluation was conducted using two prompt settings, both leveraging In-Context Learning (ICL). \texttt{Prompt 1} provides general examples as few-shot demonstrations, while \texttt{Prompt 2} supplies structured, category-specific question templates (Table~\ref{tab:prompts}).

\setlength{\tabcolsep}{2.5pt}
\begin{table}[htbp]
\tiny
\caption{Routing performance percentages of LLM routers.}
\label{tab:router-eval-results}
\centering
\begin{tabular}{lcccc}

\toprule
\textbf{Category} & \multicolumn{2}{c}{\textbf{llama3.1:8b}} & \multicolumn{2}{c}{\textbf{gpt-oss:20b}} \\
\cmidrule(lr){2-3} \cmidrule(lr){4-5}
& \textbf{\% Prompt 1 } 
& \textbf{\% Prompt 2} 
& \textbf{\% Prompt 1} 
& \textbf{\% Prompt 2} \\
\midrule
OBJECT\_ATTRIBUTES & 93.69 & 100.00 & 100.00 & 99.03 \\
PATTERN\_RECOGNITION & 100.00 & 99.61 & 100.00 & 100.00 \\
OBJECT\_COUNTING & 96.52 & 99.52 & 100.00 & 100.00 \\
ADDITION & 87.93 & 100.00 & 99.03 & 98.08 \\
SUBTRACTION & 81.97 & 100.00 & 100.00 & 100.00 \\
MULTIPLICATION & 95.83 & 100.00 & 100.00 & 100.00 \\
DIVISION & 100.00 & 100.00 & 100.00 & 100.00 \\
COMPARISON & 40.00 & 100.00 & 100.00 & 98.77 \\
\bottomrule
\end{tabular}
\end{table}

\texttt{gpt-oss:20b} achieves near-perfect performance across all categories, largely independent of prompt design. In contrast, \texttt{llama3.1:8b} is highly prompt-sensitive, with notable degradation under Prompt 1 (e.g., 40.00\% in COMPARISON), but improves to near-perfect accuracy under Prompt 2. This indicates that, with optimized prompting, the smaller model can approach the performance of the larger baseline.

\subsection{Intermediate Representation Generation Results}
The performance evaluation of the Intermediate representation generation module is evaluated using the \texttt{F1} score across three problem categories: Pattern Recognition, Object Counting, and Object Attribute Comparison (see Table~\ref{tab:ir-eval-results}).


\setlength{\tabcolsep}{4pt}
\begin{table}[htbp]
\tiny
\caption{F1-score for each problem category across models.}
\label{tab:ir-eval-results}
\centering
\begin{tabular}{lccc}
\toprule
\textbf{Model} & \textbf{Pattern Rec.} & \textbf{Object Co.} & \textbf{Attr. Comp.} \\
\midrule
gemma:7b     & 64.10\% & 67.90\% & 92.60\% \\
gpt-oss:20b  & 96.30\% & 94.50\% & 94.30\% \\
llama3.1:8b  & 84.30\% & 91.80\% & 31.40\% \\
llava:13b    & 67.70\% & 88.90\% & 63.10\% \\
\bottomrule
\end{tabular}
\end{table}

The results show that \texttt{gpt-oss:20b} achieves consistently high performance across all categories F$_1$-score $> 0.94$. In contrast, smaller models exhibit domain-specific variation: \texttt{llama3.1:8b} excels in Pattern Recognition and Counting but underperforms in Attribute Comparison, while \texttt{gemma:7b} shows the reverse trend. The multimodal \texttt{llava:13b} model demonstrates moderate but less stable performance overall.

\subsection{Automatic Asset-Level Results}
Table~\ref{tab:asset-results} shows a quantitative comparison between the two asset fallback approaches: web-based retrieval pipeline (Approach~A) and the generative synthesis pipeline (Approach~B).

\begin{table}[htbp]
\tiny
\caption{Comparison of asset generation approaches.}
\label{tab:asset-results}
\centering
\begin{tabular}{lcc}
\toprule
\textbf{Metric} & \textbf{Approach A} & \textbf{Approach B} \\
\midrule
CLIP (mean) & 0.3171 & 0.3112 \\
ImageReward (mean) & 0.6974 & 0.9665 \\
TIFA (macro average) & 0.6440 & 0.7170 \\
\bottomrule
\end{tabular}
\end{table}

Although the web-based method achieved a slightly higher \texttt{CLIP} score, the generative approach performed better on \texttt{ImageReward} and \texttt{TIFA}. This suggests that the synthesized assets were more visually coherent and more faithful to worksheet-relevant constraints such as single-object cleanliness and reduced background clutter. For the target application, these properties are important because downstream symbolic composition amplifies even small asset imperfections when icons are repeated for counting, grouping, or pattern rendering.

\section{Discussion}
The current findings reinforce the broader design choice of this work. End-to-end generation may offer broader visual coverage, but pedagogically reliable math visuals benefit from explicit control over counts, grouping, pattern structure, and highlighted attributes. By combining symbolic scene construction with open-world asset fallback, the framework improves scalability while preserving scene-level correctness.

Several limitations remain, including the current evaluation scale and reliance on repeated LLM calls, which may affect generalizability, latency, and deployment cost.

\section{Conclusion and Future Work}

This paper presented \texttt{Math2Visual-X}, a modular symbolic framework for generating pedagogically aligned visuals for lower-primary mathematical word problems. The framework extends \texttt{Math2Visual} by supporting pattern recognition, object counting, and object attribute comparison through an LLM-based routing mechanism and an open-world SVG asset fallback pipeline. Component-wise evaluation demonstrated strong routing and intermediate representation generation performance, while human evaluation showed that \texttt{Math2Visual-X} outperformed \texttt{SDXL}, \texttt{Nano Banana}, and \texttt{GPT Image} across accuracy, completeness, clarity, and cognitive load. These findings demonstrate the effectiveness of combining symbolic rendering with controlled generative asset acquisition for scalable educational visual generation.

Future work will extend support to additional lower-primary problem types, expand the evaluation benchmark, improve renderer robustness, and optimize the pipeline to reduce repeated LLM inference and deployment latency. We also plan to evaluate the framework in classroom settings to assess its impact on students' mathematical learning.

\balance

\bibliographystyle{IEEEtranN}
{\footnotesize
\bibliography{references}
}



\end{document}